\documentclass[smallcondensed, final]{svjour3}     % onecolumn (ditto)
\smartqed  % flush right qed marks, e.g. at end of proof
\usepackage{graphicx}
\usepackage{newtxtext,newtxmath} % Better alternative for times fonts
\usepackage{amsmath}
\usepackage[utf8]{inputenc} %unicode support
\usepackage[T1]{fontenc}
\usepackage{booktabs}
\usepackage{dsfont}
\usepackage{mathtools}
\usepackage{natbib}
\usepackage{nameref}
\usepackage{footmisc}
\usepackage{tabu}
\usepackage{hyperref}
\journalname{Machine Learning}

\begin{document}

\title{Feature Selection Methods for Cost-Constrained Classification in 
Random Forests}
%\subtitle{}


\author{Rudolf Jagdhuber \and Michel Lang \and Jörg Rahnenführer*}


\institute{Rudolf Jagdhuber \at
              Department of Statistics, TU Dortmund \\
              Vogelpothsweg 87\\
              44227 Dortmund\\
              \email{r.jagdhuber@gmail.com}\\
              ORCID: 0000-0002-2958-2716
           \and
           Michel Lang \at
              Institute of Statistics, LMU Munich \\
              Ludwigstraße 33\\
              80539 München\\
              \email{Michel.Lang@stat.uni-muenchen.de}\\
              ORCID: 0000-0001-9754-0393
           \and
           Jörg Rahnenführer *(Corresponding Author)\at
              Department of Statistics, TU Dortmund \\
              Vogelpothsweg 87\\
              44227 Dortmund\\
              \email{rahnenfuehrer@statistik.tu-dortmund.de}
}

\date{Received: date / Accepted: date}

\maketitle

%------------------------------------------------------------------------------%
\begin{abstract} % Max 250 words. Currently 248
Cost-sensitive feature selection describes a feature selection problem, where 
features raise individual costs for inclusion in a model. These costs allow to
incorporate disfavored aspects of features, e.g.\ failure rates of as measuring 
device, or patient harm, in the model selection process. Random Forests define 
a particularly challenging problem for feature selection, as features are 
generally entangled in an ensemble of multiple trees, which makes a post hoc 
removal of features infeasible. Feature selection methods therefore often 
either focus on simple pre-filtering methods, or require many Random Forest 
evaluations along their optimization path, which drastically increases the 
computational complexity. To solve both issues, we propose Shallow Tree 
Selection, a novel fast and multivariate feature selection method that selects 
features from small tree structures. Additionally, we also adapt three standard 
feature selection algorithms for cost-sensitive learning by introducing a
hyperparameter-controlled benefit-cost ratio criterion (BCR) for each method. 
In an extensive simulation study, we assess this criterion, and compare the 
proposed methods to multiple performance-based baseline alternatives on four 
artificial data settings and seven real-world data settings. We show that all 
methods using a hyperparameterized BCR criterion outperform the baseline 
alternatives. In a direct comparison between the proposed methods, each method 
indicates strengths in certain settings, but no one-fits-all solution exists. 
On a global average, we could identify preferable choices among our BCR based 
methods. Nevertheless, we conclude that a practical analysis should never rely 
on a single method only, but always compare different approaches to obtain the 
best results. 
\end{abstract}

%------------------------------------------------------------------------------%
\section{Background}\label{sec:int}
Cost-constrained feature selection --- often also named cost-sensitive learning 
--- describes a statistical modeling problem, where individual costs are 
assigned to candidate features. These costs may represent financial aspects, 
but can also be seen as a more general construct referring for example to 
patient harm during the sample taking process, the failure rate of a measuring
device, or a time span to raise a feature. In addition to optimizing predictive 
performance of the resulting model, cost-constrained feature selection also 
aims to control the total costs. One way to regulate the costs is to develop
flexible approaches harmonizing costs of misclassification and costs of 
features. Examples of these approaches can be found in \citet{tan1993cost},
\citet{bolon2014framework}, or \citet{zhou2016cost}. In this paper we follow a
different idea. As flexible approaches require expert knowledge to decide upon 
the final trade-off, another concept is to simply define a fixed feature cost
budget and let the feature selection algorithm find the best possible solution
within that constraint region. Research on cost-constrained feature selection 
with a fixed budget limit can for example be found in \citet{min2016semi}, who
introduce a novel semi-greedy feature selection approach, 
\citet{li2014fast, li2015fast}, who develop a randomized feature selection 
strategy focusing on computational efficiency, or \citet{jagdhuber2020cost}, 
who introduce cost-adaptations for conventional feature selection methods 
like genetic algorithms or greedy forward selection for Logistic Regression 
models. We extend this research by focusing specifically on methods tailored 
to Random Forest classification and introduce a completely new feature 
selection idea. Additionally, we also develop cost-sensitive adaptations for
existing popular methods in this context.

Random Forests \citep{breiman2001random} are an ensemble method of decision 
trees, which are constructed in a specific way that increases diversity among 
them. This includes growing each tree on a bootstrap sample of the training 
data and limiting the set of features for each internal split randomly. 
Large-scale comparisons by \citet{fernandez2014we} or
\citet{niculescu2005predicting} rate Random Forests as state-of-the-art with
respect to predictive performance. In the context of cost-sensitive feature
selection, these models define a particular challenge, as the actual 
contribution of individual features is not obvious. This makes the category of
wrapper methods for feature selection \citep{guyon2003introduction} impractical
and thus researchers often rely on univariate filter methods. These can be
simple approaches like filtering by the individual area under the receiver
operating characteristics curve (AUC)\citep{bommert2020benchmark}, but can also
be more complex measures to classify the relevance of a feature like for 
example the permutation feature importance \citep{breiman2001random}. However,
as univariate filter approaches only evaluate features one by one, relevant
multivariate aspects might be lost when reducing the dimension of the feature
space. On the contrary, more sophisticated methods like greedy sequential
forward selection, genetic algorithms \citep{holland1973genetic} or exhaustive
approaches, which all incorporate multivariate aspects to their search 
strategy, generally become computationally prohibitive with increasing data
dimensionality.

To solve these problems in a cost-constrained setup, we introduce four feature 
selection heuristics. The first is a novel algorithm named Shallow Tree 
Selection~(STS). As the name suggests, instead of selecting features directly, 
STS aims to select tree structures of one or more features. The basis for this 
selection is provided by specially fitted Random Forests. Each tree represents 
a (small) multivariate combination of features, which are considered as a 
connected entity. Unlike in feature importance filters, this property ensures 
that dependent feature combinations will never be separated during the 
selection process. From a computational performance standpoint, a big advantage 
of STS is that it does not need to repeatedly train new Random Forests models 
to evaluate candidate sets. Instead, all necessary performance components can 
be computed from the base Random Forest and its out-of-bag samples. In this
paper we describe all components of this new method and discuss technical
decisions leading to the final implementation. Besides STS, we furthermore 
propose adaptations of three common feature selection techniques (AUC 
Filtering, Permutation Feature Importance filtering and Greedy Forward
Selection) for cost-sensitive learning by introducing a specialized trade-off
metric. This metric weighs predictive performance against feature costs to
define a harmonized trade-off. It was first described in the
\textit{$\lambda$-weighted heuristic algorithm} by \citet{min2011test} and was
further assessed in \citet{min2014feature}.

To analyze the described methods, we designed an extensive simulation study
following the fundamental concepts of \citet{morris2019using}. Our main aims 
are to evaluate strengths and weaknesses of each method, to assess relevant
aspects of the trade-off metric, and to give practical recommendations for
addressing a cost-sensitive learning problem. We apply all methods on four
artificial and six real-world data sets with five pre-defined budget limits
each. Results of all proposed methods are evaluated with respect to the mean
misclassification error. All conclusions that are drawn from comparisons of
these performance measures are furthermore based on the respective Monte-Carlo
errors to account for the uncertainty introduced by the simulation setup.

Section~\ref{sec:met} starts by introducing the proposed feature selection 
heuristics. Special focus is given to our novel STS method. All main concepts 
and design decisions leading to the final implementation of STS are thoroughly
discussed in Subsection~\ref{sec:sts}. In the remaining three subsections new
cost-sensitive adaptations of AUC Filtering, Permutation Feature Importance
Filtering, and Greedy Forward Selection are introduced. Section~\ref{sec:sim}
outlines our simulation study according to the \textit{ADEMP} scheme of
\citet{morris2019using}. For a comprehensive overview of all performed 
analyses, we describe each element of \textit{ADEMP} in a separate subsection.
The results of the simulation study are then presented in Section~\ref{sec:res}
and discussed in Section~\ref{sec:dis}. In Section~\ref{sec:out}, we summarize
all findings and provide ideas for further research on this topic.

%------------------------------------------------------------------------------%
\section{Methods}\label{sec:met}
\subsection{Shallow Tree Selection (STS)}\label{sec:sts}
This section introduces the general idea of STS and thoroughly discusses all 
concepts that lead to the final method definition. While this approach includes 
many codependent elements, the general process can be crudely structured into
three main steps, which will be discussed in more detail in the following
subsections:
\begin{enumerate}
\item Generate an ensemble of trees from an adapted Random Forest approach
using all candidate features (Subsection~\ref{sec:base}).
\item In a greedy forward selection manner, iteratively select the best tree 
according to a custom trade-off criterion and add it to the result set. Stop 
the iterative procedure when the available feature cost budget is depleted
(Subsection~\ref{sec:sel}).
\item Re-fit a Random Forest with the implicitly selected feature set of the 
last ensemble (Subsection~\ref{sec:stop}).
\end{enumerate}
Section~\ref{sec:summary} finally provides a summarizing overview of the final 
method implementation.

%%%%%%%%
\subsubsection{Generating the Base Ensemble}\label{sec:base}
An optimal ensemble of trees contains a diverse set of trees from different
combinations of relevant features. The general approach of Random Forests is 
well-suited here as it creates an ensemble of heterogeneous trees by 
bootstrapping the data and limiting the feature candidates allowed for each 
split in the tree growing process. In the context of a feature cost budget, 
however, the total costs of individual trees play an important role as well. 
While Random Forests decorrelate the trees in the ensemble, they do not limit 
the number of features or the total feature costs of a single tree. When 
budgets are small, or many non-redundant features are relevant, then oftentimes
individual trees already exceed the given feature budget. To address this 
problem, trees with limited costs need to be constructed. However, the 
implementation of an adapted tree growing algorithm that holds a fixed budget 
is not trivial, and would for instance need an additional set of rules for the 
splitting order, when only a fixed set of splits fit the budget. A simple and 
ready-to-use surrogate solution is to instead define a maximum depth 
$d_\text{max}$ for each tree. Limiting the tree depth corresponds to defining 
an upper bound of distinct features per tree. While this does not allow the
definition of an exact cost budget, it is a practical strategy to scale down 
the average tree costs. The general formula for the maximum number of features 
$p$ that a tree with depth $d_\text{max}$ can include is 
$p = 2^{d_\text{max}} - 1$.

$d_\text{max}$ can be seen as a hyperparameter to regulate the trade-off of
complexity and cost. Deep trees are typically more expensive, but also have the 
ability to describe complex multivariate relations. Very shallow trees often 
come at little cost and therefore allow greater flexibility in the greedy tree 
selection steps later. Yet, they only describe small structures, which 
typically do not perform well on their own. An approach to generally solve the 
problem of choosing $d_\text{max}$ is to combine multiple base forests with 
$d_\text{max} = 1, 2, \dots, d$. The upper bound $d$ can then for example be 
set to the maximum value for sensible combinations within the budget limit. By 
including trees of all depths up to a certain level, it is now for the greedy 
forward tree selection to decide if the additional costs of a deeper tree is 
worth its added predictive value. Note that trees with $d_\text{max} = 1$ 
define a special case, where only a single tree per feature is fitted instead 
of a full Random Forest.

%%%%%%%%
\subsubsection{Greedy Forward Tree Selection}\label{sec:sel}
After growing an adequate \textit{base ensemble} of candidate trees 
$T = \{T_1, T_2, \dots, T_{n_\text{trees}}\}$, an empty 
\textit{result ensemble} $R = \emptyset$ with total cost $c_R = 0$ is set up. 
The goal of the subsequent greedy forward selection is to iteratively identify 
the most suited tree from the base ensemble $T$ and add it to the result 
ensemble $R$. The term \textit{most suited} is defined by a trade-off criterion 
that is introduced in the following paragraph. 

\paragraph{Benefit-Cost Ratio Selection Criterion.}\label{sec:bcr}
In cost-constrained feature selection problems there are two relevant aspects 
that define the quality of an ensemble. The first aspect is predictive 
performance. In the context of Random Forests, predictive performance of 
a \textit{single} tree can conveniently be evaluated by computing the mean 
misclassification error (MMCE) on the out-of-bag (OOB) samples of the tree. 
However, to compute the \textit{combined} MMCE of an ensemble of trees, it is 
important not to simply average the individual MMCEs of all trees. This becomes 
more clear if we for example assume three trees, which classify the same 
single observation and two of them make a correct prediction, while one 
misclassifies it. Here, the average of the individual MMCEs is $\frac13$, yet 
the ensemble itself makes a correct classification and thus has an actual error 
of $0$. To correctly estimate the MMCE, each tree needs to cast a vote for its 
out-of-bag samples and a majority vote decides in the end on the final 
classification of each sample. As a practical decision here, we assume samples 
without a vote to be falsely classified. While this generally results in a too 
conservative MMCE estimate, it encourages the algorithm to pick trees from 
different OOB samples to reduce the overall error. The second aspect that
defines the quality of a tree ensemble is its total feature cost. This value
simply computes as the sum of all costs of included features. In accordance to
the generally used definition of cost-sensitive learning, we assume that 
feature costs are only assigned once per feature and an already `paid' feature
can thus be reused without additional costs. 

To address both aspects in the greedy forward selection, a trade-off criterion 
is introduced. The benefit-cost ratio for STS divides the decrease in MMCE of 
the ensemble $R$ when adding a tree $T_i$ by the additional cost that $T_i$ 
generates. This allows a combined criterion even though both measures live on 
different scales. To guide the relative importance of each of these aspects, a 
hyperparameter $\xi$ is introduced. The resulting measure is given by

\begin{equation}\label{eq:bcr}
  \text{BCR}_\text{STS}(T_i, R, \xi) = 
  \frac{\text{MMCE}_\text{OOB}(\{R \cup T_i\}) - 
  \text{MMCE}_\text{OOB}(\{R\})}
  {\left(c_{\{R\, \cup\, T_i\}} - c_{\{R\}}\right)^\xi}.
\end{equation}
with $\text{MMCE}(\emptyset) = 0.5$ and $c_\emptyset = 0$. The BCR can be used 
to evaluate the impact of adding different candidate trees to an ensemble. The 
optimal value of this measure is its minimum.

\paragraph{Selecting the Hyperparameter $\xi$}\label{sec:hyper}
The hyperparameter $\xi$ allows to control the trade-off between cost and
performance. Our main strategy to select $\xi$ is hyperparameter tuning. In 
order to tune $\xi$, a \textbf{search space}, a \textbf{resampling strategy} 
and a \textbf{tuning method} need to be specified \citep{bischl2016mlr}. The 
search space for $\xi$ is an interval in which the optimal value is expected 
to lie. As cost is an aspect we aim to penalize, the lower limit of this 
interval can be chosen as 0. The upper limit however is technically unbounded 
and needs to be estimated for the problem at hand. The resampling strategy 
defines the data approach that is used to evaluate the different candidate 
values. Cross-validation or using a hold-out portion of the data represent 
typical choices. In the context of Random Forests, a practical strategy is to 
use the OOB samples as an independent data portion. Finally, the tuning method 
refers to the algorithm that iteratively proposes new candidates for $\xi$. A 
grid search is a simple and practical strategy for this matter. It defines a 
fixed (equidistant) grid of values within the search space. This makes it 
robust against local optima. Overall, the hyperparameter tuning strategy 
described here is comparable to the \textit{Competition Strategy}, which was 
proposed in a different context by \citet{min2011test}. A faster but more 
complex alternative is model based optimization (aka Bayesian optimization) 
\citep{jones1998efficient}, which is implemented for example in the R package 
\texttt{mlrMBO}~\citep{bischl2017mlrmbo}. This method is particularly suited 
for expensive black-box functions and provides a reasonable alternative to 
optimize $\xi$ in the STS method.

Besides hyperparameter tuning, we also evaluate alternatives that avoid the 
optimization of the hyperparameter. These can be formulated as edge cases in 
terms of $\xi$. The first such edge case is given by $\xi = 0$ and is named 
\textit{Cost-Agnostic} in the following. With this choice of $\xi$, the BCR of 
Equation~(\ref{eq:bcr}) is reduced to the decrease in MMCE. Hence, individual 
selection steps focus on MMCE only, and costs are just considered as a 
constraint regarding the overall budget limit. A second edge case is $\xi = 1$. 
This choice can be interpreted as using a simplified BCR without a 
hyperparameter, like for example found in \citet{paclik2002feature}, or 
\citet{leskovec2007cost}. Compared to the cost agnostic approach, the main
appeal of the simple BCR strategy is that it incorporates costs in the 
selection process, while still avoiding an optimization of $\xi$.

\paragraph{Candidate Trees without Costs}\label{sec:costfree}
A final aspect to consider when using a greedy forward selection approach with 
the BCR criterion of Equation~(\ref{eq:bcr}) is that trees can become cost-free.
That means that at some point, there are candidate trees that only consist of 
features already present in $R$. Then the denominator of the BCR criterion  
becomes $0$. To avoid this, a decision on how to handle possible cost-free 
trees needs to be made at the end of each iteration. One option is to add all 
such trees automatically to the result ensemble. The downside of this approach 
is that adding a possibly large number of trees at one iteration decreases the 
impact of all following iterations. For example, the predictions of every 
single candidate tree to consider in the second iteration may already be 
outvoted by hundreds of votes from identically structured trees that were added
in the first iteration. A further aspect is, that just because a single tree
structure is identified to be useful, it does not mean that all existing
combinations of those features are.

The alternative option is to remove all cost-free trees from the candidate pool 
at the end of each iteration. The downside of this approach is that only one 
structure of each feature combination can be added to the result ensemble.
However, as this aspect only has minor effects on the final result and 
decreasing the pool of candidates furthermore decreases the computational 
complexity of the algorithm, the latter option was chosen for the STS method.

%%%%%%%%
\subsubsection{Feature Selection Result}\label{sec:stop}
We continue to add trees from the base ensemble $T$ to the result ensemble $R$,
until either $T$ does not contain any further candidate trees, or none of these 
trees fits the remaining feature cost budget. A specific design aspect of STS 
is that the algorithm does not aim to build up a highly predictive result
ensemble that may then be used as a model. Instead, it only aims to identify
well-matching sets of features. Using the ensemble itself as a prediction model
would have multiple disadvantages. First, the trees used for STS have limited
depth and thus typically cannot compete with their fully grown counterparts.
Second, the final ensemble often contains relatively few trees, even if 
all cost-free trees would be added in the end. As predictive performance of
Random Forests increases with the number of trees, the forests constructed in
Subsection~\ref{sec:sel} generally show relatively poor performance. Therefore,
an essential aspect of STS is that it does not focus on the explicit result
ensemble itself, but only considers the implicitly selected set of features
contained in the trees of the ensemble.

To create the final prediction model, a new Random Forest is fitted from the
implicit feature set of the result ensemble. Note that no early stopping of 
the greedy forward selection is applied. Therefore, the final feature set may
still contain a large number of noise features, especially in situations with
high budget limits. However, for Random Forests, which usually just ignore 
noise features, this does not pose a problem.

%%%%%%%%
\subsubsection{Summary}\label{sec:summary}
\begin{figure}[ht]
\includegraphics[width=\textwidth]{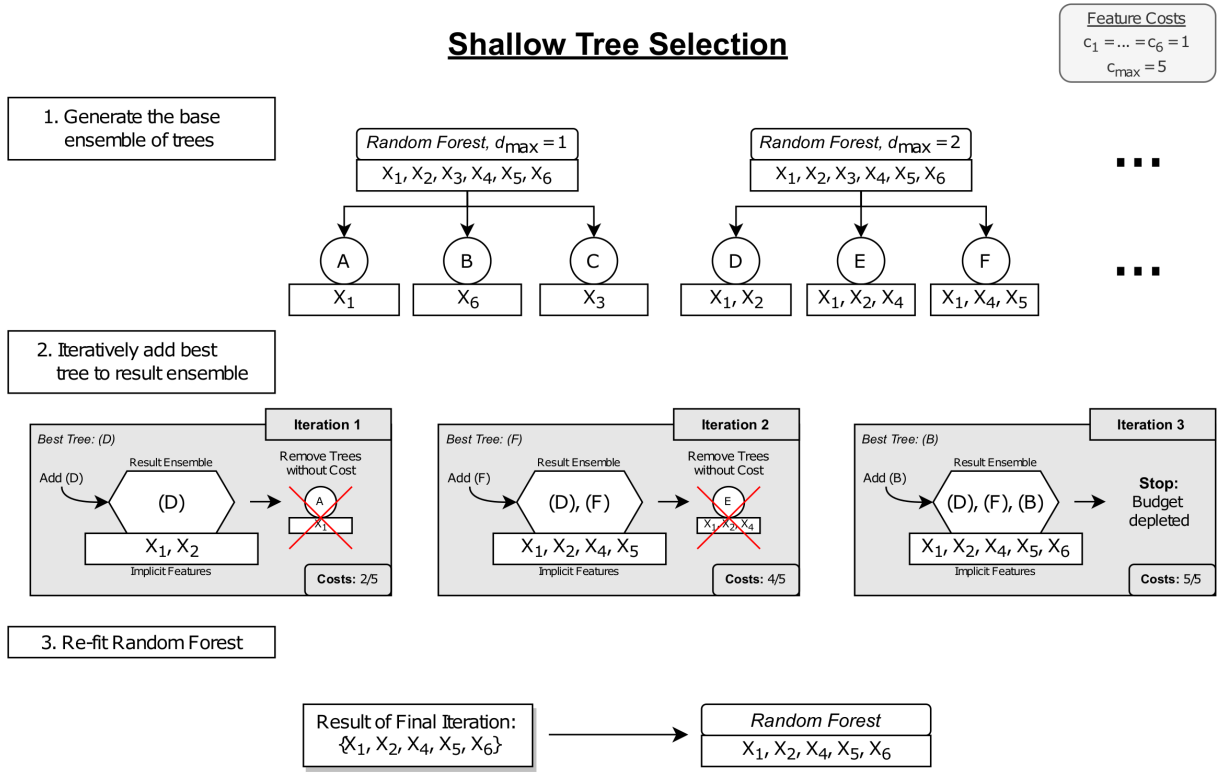}
\caption{Schematic example of the Shallow Tree Selection method with six 
candidate features. In step 1, Random Forests with a maximum tree depth of one 
and two are fitted. Each forest generates three trees for a total of six 
candidate trees labeled (A) to (F) (cf. Subsection~\ref{sec:base}). Step 2
describes the greedy forward selection, which starts with an empty result
ensemble. In the first iteration the current best candidate tree (D) is added 
to this result ensemble, which therefore now implicitly contains the features
$X_1$ and $X_2$. `Best' is defined by means of the BCR criterion (cf.
Subsection~\ref{sec:bcr}). After that, candidate tree (A) only contains features
that are already present in the result ensemble. It is thus removed from the 
list of candidate trees (cf. Subsection~\ref{sec:costfree}). In the second 
iteration, the most suited tree is (F). (F) is added to the result ensemble, 
which analogously to the first iteration leads to the removal of the now 
cost-free tree (E). The final iteration adds tree (B) and fills up the budget.
This concludes the greedy forward selection. Step 3 uses the implicit feature
set of the last iteration of the greedy forward selection to fit a new Random
Forest with it. This step is important to overcome weaknesses, which result for
instance from the limited tree depths (cf.
Subsection~\ref{sec:stop}).\label{fig:sts-scheme}}
\end{figure}

This section summarizes all previously introduced design elements and presents 
a general method overview. A simple working example is given in 
Figure~\ref{fig:sts-scheme} to illustrate the main principles of STS. The 
caption of this figure gives a thorough description of each step and refers to
relevant subsections for further information. 

%------------------------------------------------------------------------------%
\subsection{Feature Filtering by AUC}\label{sec:met-auc}
To broaden the field of cost-sensitive feature selection methods in Random 
Forests, we also introduce budget-constraint adaptations for commonly used 
standard approaches in this context. Because of the computational complexity of 
this setup, filter methods based for example on Decision Stumps, the Gini-index,
or the univariate \textbf{A}rea \textbf{U}nder the receiver operating 
characteristic \textbf{C}urve~(AUC)\citep{bommert2020benchmark} are typical 
choices for feature selection. These methods individually evaluate the 
univariate ability of each single feature to separate the response classes. For
this paper, we introduce a cost-sensitive adaptation of the AUC filter method.
The AUC is a popular trade-off measure, which incorporates both sensitivity and 
specificity. This makes it a very common choice for example in diagnostic 
applications, or in situations with imbalanced classes. For each feature $X_j$, 
we use the following prediction rule of the binary response variable 
$Y$: $\hat{Y} = I_{[c,\infty)}(X_j)$, with $I$ denoting the indicator function. 
That is, if $X_j$ is greater than a certain threshold $c$, we predict class 1. 
Otherwise, class 0 is predicted. The Receiver Operating Characteristic curve 
displays the sensitivity and specificity of this classification rule for all 
choices of $c$. The area under this curve defines a trade-off measure of both 
aspects and is hence often used to assess the overall quality of a prediction 
rule. For this measure, a value of $0.5$ corresponds to a random prediction and 
values of 0 and 1 denote a perfect separation of the class variable. We define 
a monotonous version of the AUC by
\begin{equation} \label{eq:auc}
  J_\text{AUC}(X_j) = 2\,|\text{AUC}_{\hat{Y}}(X_j) - 0.5|,
\end{equation}
which transforms the value of a random prediction to 0 and a perfect separation 
to 1. To introduce a cost-sensitive version of this measure, a benefit-cost 
ratio approach similar to Equation~(\ref{eq:bcr}) is used. For each feature
$X_j$ with feature cost $c_{X_j}$, the benefit-cost ratio measure for AUC
filtering is given by
\begin{equation} \label{eq:bcrauc}
  \text{BCR}_\text{AUC}(X_j, \xi) = \frac{J_\text{AUC}(X_j)}
  {c_{X_j}^\xi},
\end{equation}
where $\xi$ is a hyperparameter to control the trade-off between performance and
cost. We compare different strategies for this hyperparameter in the simulation
study in Section~\ref{sec:sim}.

After obtaining an individual BCR value for each feature, a top-down approach 
is used to built up a feature set that meets the budget limit. Features are 
added to the result set in order of their $\text{BCR}_\text{AUC}$ rank, but 
only, if the cost of the resulting model does not exceed the budget. The 
process is stopped, when the cost of any remaining feature would exceed the 
budget \citep{jagdhuber2020cost}.

%------------------------------------------------------------------------------%
\subsection{Feature Filtering by Permutation Feature Importance (pFI)}\label{sec:met-pfi}
Another popular approach particular to Random Forests is the 
\textbf{p}ermutation \textbf{F}eature \textbf{I}mportance (pFI) 
\citep{breiman2001random}. It is a heuristic measure, which assesses the 
relevance of individual features in a full Random Forest model. To calculate 
this measure, the observations of a single feature $X_j$ are permuted and the 
OOB error of the resulting Random Forest is computed. As the permutation 
removes any predictive information of the variable $X_j$, the difference of this 
OOB error to the original one describes a measure of importance for $X_j$. This 
measure has both univariate and multivariate aspects. On the one hand, the score 
of a feature is based on the multivariate context of a Random Forest model and 
could thus be labeled multivariate. On the one hand, each feature is assigned an 
individual score, which is then typically used for a one-by-one feature 
selection. As relevant combinations may be separated, a high importance value 
does not necessarily imply a good performance in a smaller subset of features. 

We label the permutation feature importance measure $J_\text{pFI}(X_j)$. It can 
take values between 0 and 1, with larger values corresponding to a higher 
importance of the respective feature. Similar to the AUC, we propose a 
cost-sensitive version of this measure with
\begin{equation} \label{eq:4-bcrpfi}
  \text{BCR}_\text{pFI}(X_j, \xi) = \frac{J_\text{pFI}(X_j)}{c_{X_j}^\xi},
\end{equation}
where $\xi$ is a hyperparameter to control the trade-off of performance and 
cost. To built up a final result set from the individual feature assessments, 
the same approach as described in Subsection~\ref{sec:met-auc} is used.

%------------------------------------------------------------------------------%
\subsection{Random Forest Forward Selection}\label{sec:met-fs}
The third additional approach of this paper introduces a common feature 
selection strategy found for example in linear models for the context of Random 
Forests. A greedy sequential \textbf{F}orward \textbf{S}election (FS) starts 
with an empty set of features and iteratively adds the one feature that best 
extends the current result set. \textit{Best} is typically defined by means of 
goodness of fit, or predictive performance. In our context, this translates to 
building a Random Forest model for each candidate set and assessing the 
respective OOB errors. The central problem of this approach is the high 
computational complexity of Random Forests in general associated with the large 
number of models needed. Consider a setting with 1000 features. In the first 
iteration, the forward selection grows and assesses 1000 (one feature) Random 
Forests. In the second iteration there are 999 Random Forests to be evaluated. 
After 100 iterations, a total of 95050 models needs to be computed. The only 
exception to this would be situations with small budgets, where many paths 
could be eliminated early on. For all other problems, however, the 
computational effort of this method can be immense. Even with fast
implementations, like for example found in the R package
\texttt{ranger}~\citep{wright2015ranger}, the overall run-time is hard to 
manage in practice. 

Nevertheless, we implemented this approach for our analyses in a cost-constraint
context. At iteration $k$, the forward selection needs to decide which feature 
$X_j$ is added to the current result set $S$. We propose the following 
cost-sensitive criterion for this decision:
\begin{equation}\label{eq:bcrfs}
  \text{BCR}_\text{FS}(X_j, S_k, \xi) = \frac{\text{MMCE}_\text{OOB}\!
  \big(\text{RandomForest}(\{S_k \cup X_j\})\big) -
  \text{MMCE}_\text{OOB}\!\big(\text{RandomForest}(\{S_k\})\big)}
  {\big(c_{\{S_k\, \cup\, X_j\}} - c_{\{S_k\}}\big)^\xi}
\end{equation}
The stopping criterion and the decision on a final solution follow the ideas 
discussed in Subsection~\ref{sec:stop}. Similar to the previous sections,
Equation~(\ref{eq:bcrfs}) introduces a hyperparameter $\xi$ to control the
trade-off of cost and performance. Tuning this hyperparameter requires to 
execute the feature selection process for a reasonable set of candidate values. 
With the immense run-time of each FS run, this was not feasible for the 
simulation setup of Section~\ref{sec:sim}. Our most promising attempt used a 
model-based optimization approach, which however still required approximately 
10 hours for the tuning process of one simulation run. Therefore, we do not 
consider hyperparameter tuning for FS and only assess the cost-agnostic and the
simple BCR strategy in the following.

%------------------------------------------------------------------------------%
\section{Simulation Study}\label{sec:sim}
The structure of this section follows \textit{ADEMP}, a framework proposed by 
\citet{morris2019using} to systematically define simulation studies. ADEMP is 
an acronym for: \textbf{A}ims, \textbf{D}ata-generating mechanisms,
\textbf{E}stimands, \textbf{M}ethods, \textbf{P}erformance measures. In the 
following, we address each element of ADEMP in an individual section to provide
a thorough outline of all performed analyses.

\subsection{Aims}\label{sec:aims}
Our simulation study evaluates feature selection methods for Random Forest 
classification tasks in a cost-sensitive setup. We do not intend to 
claim a general best method for all situations, but rather to understand 
strengths and weaknesses of the analyzed methods in realistic data settings. 
Additionally, we compare different strategies for our BCR criterion to assess 
the impact of a hyperparameter controlled trade-off approach compared to more 
simple strategies. In summary, the simulation study addresses the following 
main questions:
\begin{enumerate}
  \item Is there a general benefit from using a simple BCR approach ($\xi = 1$)
  in the proposed methods instead of a cost agnostic strategy ($\xi = 0$)? If 
  not, is there any other globally best setting for $\xi$?
  \item Does the hyperparameter tuning approach for $\xi$ generate a relevant 
  improvement compared to the alternatives discussed in Aim~1?
  \item Which of the analyzed methods show strengths/weaknesses for data
  simulated with only univariate effects on a binary response variable? Which
  do so for multivariate effects?
  \item Does the extent of the budget constraint $c_\text{max}$, or a
  correlation between cost and relevance of a feature affect these 
  results?
  \item Can we observe similar or further effects when applying the proposed
  methods on real-world data? 
  \item Can the proposed methods outperform their respective cost-agnostic
  baseline versions on real data?
  \item Can we categorize strengths and weaknesses of each method by the known
  meta information of the data sets?
\end{enumerate}
After presenting the simulation results in Section~\ref{sec:res}, a detailed
step-by-step discussion of each aim is provided in Section~\ref{sec:dis}.

\subsection{Data-generating mechanism}
The simulation study is divided into two parts. The first part analyzes four 
settings on artificially generated data, while the second part evaluates six 
real-world data sets. In this section, we describe the processes to generate 
all artificial data sets and provide background information on the origins of 
the real-world data sets.

\subsubsection{Artificial Data Settings}
For the artificial simulation settings, we consider data sets of 
$n_\text{obs} = 500$ observations with $p = 200$ continuous numeric features, 
representing a typical set-up for example in metabolomic studies or phase-III 
trials \citep{banas2018identification}. Data sets are generated following the 
framework of \citet{boulesteix2017ipf}. First $n_\text{obs}$ realizations of 
the binary response variable are drawn from the Bernoulli distribution 
$\mathcal{B}$(0.5). After that, the corresponding features to each response 
value are drawn from the $p$-dimensional normal distributions
\begin{align}\label{eq:mvt}
  \begin{split}
X_1,\dots , X_p\ |\ Y=1 &\sim\mathcal{N}_p(\boldsymbol\mu,\boldsymbol\Sigma),\\
X_1,\dots , X_p\ |\ Y=0 &\sim\mathcal{N}_p(\mathbf{0}_p,\boldsymbol\Sigma) ,
  \end{split}
\end{align}
with mean vector $\boldsymbol\mu$ defined as
\begin{equation}\label{eq:mudef}
\boldsymbol\mu^T = (\underbrace{\beta_1,\dots ,\beta_{p_\text{rel}},
0,\dots,0}_{p}).
\end{equation}
To include a variety of weak, mediocre and strong effects, we draw the
$p_\text{rel}=100$ realizations of $\beta_j, j = 1, \dots, p_\text{rel}$ from a 
truncated normal distribution $\mathcal{N}_{[-1, 1]}(0, 0.5)$. This truncated 
normal distribution can be interpreted as a usual normal distribution, where 
only realizations within the interval $[-1, 1]$ are considered valid. We thus 
avoid unrealistically dominant effects. The remaining elements of 
$\boldsymbol\mu$ are set to zero, and the corresponding features can be 
interpreted as noise.

For the covariance matrix $\boldsymbol\Sigma$, we consider two alternatives.
The first choice represents a scenario of independent features with only
univariate relations to the response variable. This assumes 
$\boldsymbol\Sigma = \boldsymbol I_p$ and is labeled ``Independent Data''.
In many real-world applications, however, the assumption of completely 
independent features does not hold true \citep{de2008nmr}. Therefore, we also
consider a scenario with ``Correlated Data'', i.e.\ a non-diagonal covariance
matrix $\boldsymbol\Sigma$. More specifically, we assume $20$ groups of 
mutually correlated features, which corresponds to the following block-diagonal
covariance matrix:
\begin{equation}\label{eq:blocksig}
\boldsymbol\Sigma = \begin{pmatrix}
  \boldsymbol{S}(\rho_1) & \mathbf{0} & \mathbf{0} \\
  \mathbf{0} & \ddots & \mathbf{0}\\
  \mathbf{0} & \mathbf{0} & \boldsymbol{S}(\rho_{20})\\
\end{pmatrix},
\end{equation}
where $\boldsymbol{S}(\rho_i), i = 1, \dots, 20$ is a 
$(\frac{p}{20}\times \frac{p}{20})$ matrix with ones on the diagonal and 
$\rho_i$ outside the diagonal. Realizations of $\rho_i$ are drawn from the
uniform distribution on the interval $[0, 1]$. By only considering positive
values for $\rho_i$, we can ensure that $\boldsymbol\Sigma$ is positive
definite. To randomly distribute informative features (i.e.\ the non-zero
entries of $\boldsymbol\mu$) and noise features across this correlation
structure, the order of the features in this matrix is permuted, permuting
columns and rows of $\Sigma$ at the same time, before drawing from
$\mathcal{N}_p(\cdot, \Sigma)$. One goal of defining a non-diagonal covariance
matrix is to create a multivariate modelling problem. It may not seem obvious
that correlated features automatically correspond to multivariate effects on 
the response variable. However, with the data-generating mechanism of
Equation~(\ref{eq:mvt}) any pair of correlated features with non-identical
effect sizes induces an additional separation that is proportional to the 
extent of the correlation (cf. Figure~2 of \citet{jagdhuber2020cost}). Note 
that this consequently also changes the interpretation of the defined effect
sizes in Equation~(\ref{eq:mudef}). In these settings, the absolute size of
$\beta$ does not define the relevance of the respective feature anymore and
therefore a feature with effect size zero can no longer generally be seen as
noise. However, as we avoid interpretations concerning, e.g., the number of
detected relevant and noise features, this does not affect the analyses in the
following.

For the feature costs $c_j, j=1,\dots,p$, we also consider two alternatives. In 
the first scenario (labeled ``Independent Costs''), feature costs are drawn 
from a uniform distribution on the interval $[0.1, 1]$. This assumes that 
costs $c_j$ are independent of the respective effect sizes $\beta_j$. In 
practice, it can also be plausible to assume that more valuable features might
be more expensive on average. We therefore consider a scenario of ``Correlated
Costs'' as well. In this scenario, the cost of the $j$-th feature is computed
using the absolute value of $\beta_j$ and adding normal distributed uncertainty
to it. After that, the value is truncated to our intended cost range 
$[0.1, 1]$. Formally this can be written as 
\begin{equation}\label{eq:corcost}
c_j = \min(1, \max(0.1, |\beta_j| + \epsilon_j)),
\end{equation}
where $\epsilon_j\sim\mathcal{N}(0, 0.2)$. 
For the overall budget limit $c_\text{max}$ we consider five values, which
represent small, medium and large budgets, by defining 
$c_\text{max} \in \{1, 2, 5, 10, 30\}$. 

Altogether, two scenarios for the choice of $\boldsymbol\Sigma$ and two
scenarios for feature cost correlations are varied and each setting is 
evaluated at five feature cost budget limits. To thoroughly analyze all 
possible effects, we use a factorial design resulting in four settings named A
to D with five budget limits each. An overview of these settings is given in
Table~\ref{tab:settings}.
\begin{table}[ht]\centering
\begin{tabular*}{\textwidth}{@{\extracolsep{\fill}}llll}
\toprule
& Data Covariance $\boldsymbol\Sigma$ & Relationship $\beta_i\sim c_i$ & Cost budget limit $c_\text{max}$\\
\midrule
Setting A & Independent Data & Independent Costs & \{1, 2, 5, 10, 30\}\\
Setting B & Independent Data & Correlated Costs & \{1, 2, 5, 10, 30\} \\
Setting C & Correlated Data & Independent Costs & \{1, 2, 5, 10, 30\} \\
Setting D & Correlated Data & Correlated Costs & \{1, 2, 5, 10, 30\} \\
\bottomrule
\end{tabular*}
\caption{Overview of the factorial design of the artificial data simulations
varying the covariance matrix $\boldsymbol\Sigma$, the dependency of costs to 
effect sizes and the feature cost budget limit 
$c_\text{max}$.\label{tab:settings}}
\end{table}

All methods are applied on a set of $n_\text{sim} = 100$ training data sets, 
which are drawn as follows: In a first step, a fixed value for 
$\boldsymbol\mu$, which is used in all settings A to D, and the block-diagonal 
matrix $\boldsymbol\Sigma$ of Definition~\ref{eq:blocksig}, which is used for 
settings C and D, are drawn. Furthermore, a set of independent costs for
settings A and C and a set of correlated costs for setting B and D are
generated. With these elements, $n_\text{sim} = 100$ training data sets with 
$n_\text{obs} = 500$ observations and one test data set with 
$n_\text{test} = 5000$ observations are drawn respectively for the independent
and correlated settings according to the framework described above in this
section. The analysis of these data sets is structured into the following steps.
All methods are applied to the training data sets for a grid of values $\xi$ 
including $\xi = 0$ and $\xi = 1$. Among these grid values, the resulting 
feature sets at $\xi=0$ (cost-agnostic results), $\xi=1$ (simple BCR results) 
as well as the set with the lowest OOB error (hyperparameter tuning results) 
are selected for every feature selection method. These sets are used to build 
full Random Forest models on the training data, which are then evaluated on the 
test data set to obtain the MMCE.

\subsubsection{Real-World Data Settings}
In addition to the artificial simulations, we also include a real-world 
study in our analyses. The online platform Open-ML \citep{OpenML2013} provides 
a large collection of well documented data sets that can be filtered by 
relevant meta-information like the number and type of features, the response 
variable type, the amount of missing values and more. For the objectives of 
this paper, six binary classification data sets met all requirements, which 
included a sample size of at least 500 observations, no missing values, and a 
total between 30 and 500 non-categorical features. We aimed to select data from 
many different fields to cover a large variety of practical applications. The 
exception to this is image analysis, as a setup of different feature costs for 
individual pixels seems unreasonable. An overview of all real-world data sets 
used in this paper is given in Table~\ref{tab:realsets}. A short introduction 
on every data set is provided in the following.
\medskip

\begin{table}[ht]\centering
\begin{tabular*}{\textwidth}{@{\extracolsep{\fill}}llllll}
\toprule
Data Set & $p$ & $n_\text{obs}$ & $P(y=1)$ & Field & Source\\
\midrule
Ada & 48 & 4562 & 0.248 & Social Studies &
\href{https://www.openml.org/d/40993}{https://www.openml.org/d/40993}\\
Author & 70 & $\;$ 841 & 0.377 & Text Mining & 
\href{https://www.openml.org/d/458}{https://www.openml.org/d/458}\\
Qsar & 41 & 1055 & 0.337 & Biology & 
\href{https://www.openml.org/d/1494}{https://www.openml.org/d/1494}\\
Spam & 57 & 4601 & 0.394 & Text Mining & 
\href{https://www.openml.org/d/44}{https://www.openml.org/d/44}\\
Tokyo & 44 & $\;$ 959 & 0.639 & Server Performance & 
\href{https://www.openml.org/d/40705}{https://www.openml.org/d/40705}\\
Wdbc & 30 & $\;$ 569 & 0.373 & Medicine & 
\href{https://www.openml.org/d/1510}{https://www.openml.org/d/1510}\\
\bottomrule
\end{tabular*}
\caption{Overview of all real-world data sets used in the simulation study. For
further details see the data pages on Open-ML and the provided 
references.\label{tab:realsets}}
\end{table}

\noindent
\textit{Ada.} The main task of ADA is to identify citizens with high revenue.
This is a two-class classification problem. The raw data from the census bureau
is also known as the ``Adult'' database in the UCI Machine-Learning Repository
\citep{dua2019}. The 14 original features include age, work-class, education,
marital status, etc., but are pre-processed and scrambled into a 48 feature
continuous numeric representation \citep{openMLada}.
\smallskip

\noindent
\textit{Author.} The original version of this data set is part of a 
collection of data sets used in the book ``Analyzing Categorical Data'' by 
\citet{simonoff2013analyzing}. The categorical response variable consists of the
author labels ``Austen'', ``London'', ``Milton'' and ``Shakespeare''. Features 
correspond to word frequencies of 70 different words. The data set used in this 
paper is a binarized version of the original set, where the response variable is
converted by re-labeling the majority class (``Milton'') as $1$ and all others
as $0$.
\smallskip

\noindent
\textit{Qsar.} The QSAR biodegradation data set
\citet{mansouri2013quantitative} was built in the Milano Chemometrics and QSAR 
Research Group. The data have been used to develop QSAR (Quantitative Structure 
Activity Relationships) models for the study of the relationships between 
chemical structure and biodegradation of molecules. A total of 1055 chemicals 
result in 41 numeric molecular descriptor features and one binary response 
variable with classes \textit{ready biodegradable} and 
\textit{not ready biodegradable}. 
\smallskip

\noindent
\textit{Spam.} A data set of Hewlett-Packard Labs containing 4601 E-mails 
together with a user based classification into ``Spam'', or ``No Spam''. 
Features include word and character frequencies as well as further general 
numeric measures on the text composition \citep{dua2019}. 
\smallskip

\noindent
\textit{Tokyo.} Performance-Co-Pilot (PCP) data for the Tokyo server at 
Silicon Graphics International. The data characterizes the server performance 
as either ``good'' (1) or ``bad'' (0). The 44 numeric features include technical
measures like the average percentage of CPU time spent for user code, read and 
write rates, or average free memory. The instances are measurements generated 
by the PCP software every five seconds. 
\smallskip

\noindent
\textit{Wdbc.} The Wisconsin Diagnostic Breast Cancer data set describes 
characteristics of cell nuclei from a digitized fine needle aspiration 
procedure (FNA) of a breast mass. Ten real-valued features are computed for 
each of three cell nuclei, yielding a total of 30 descriptive features. These 
features include for example radius, smoothness, and symmetry. The response 
variable records the prognosis  as either ``benign'' (0) or ``malignant'' (1).
\medskip

In their original form, none of these real-world data sets includes
feature costs. Therefore, a cost setting analogous to the ``Independent Costs'' 
scenario of the artificial data is simulated. As the true effects of features 
are not simulated and estimations of these require strong and possibly 
unrealistic assumptions, we do not generate a ``Correlated Costs'' scenario 
here. Instead, all feature costs are drawn from the uniform distribution on the 
interval $[0.1, 1]$. To also analyze possible effects of this random draw, for
the Spam data set, we define five different cost vectors and compare the
obtained results. As the number of relevant and total features varies between
data sets, the feature cost budget limits are chosen individually. For each 
data set, five candidates representing a range of small to very large
constraints are analyzed. 

Furthermore, the real-world data sets are not split into training and testing 
sets. Thus, we create these splits manually. $n_\text{sim} = 100$ random split 
points, which partition data sets into $\frac23$ of observations to be used as 
training sets and $\frac13$ of observations to be used as testing set are
defined. The analysis itself is structured similarly to the artificial 
settings. All methods are applied for a grid of $\xi$ values to each of the
training partitions. Among these grid values, the resulting feature set at
$\xi=0$ (cost-agnostic baseline) and the set with the lowest OOB error
(hyperparameter tuning results) are selected. These sets are used to build full
Random Forest models on the training partitions, which are then evaluated on 
the test partitions to obtain the MMCE.

\subsection{Estimands and Targets}
Our statistical analysis task is to select a feature subset from a pool of 
candidate features and then predict a binary response variable with a Random 
Forest model. First, the feature selection and Random Forest model fit are
performed on training data. This model then predicts the response variable of 
an independent test data set. In the sense of \citet{morris2019using}, this 
response variable represents our final estimand and the respective predictions
of each candidate model can be seen as the target of the analysis.

\subsection{Methods}
The simulation study analyzes methods for feature selection in binary Random 
Forest classification problems. The compared methods are:
\medskip

\begin{tabular}{ll}
\textbf{STS}:& Shallow Tree Selection (see section~\ref{sec:sts}),\\
\textbf{AUC}:& AUC Filter (see section~\ref{sec:met-auc}),\\
\textbf{pFI}:& Permutation Feature Importance (see section~\ref{sec:met-pfi}),\\
\textbf{FS-0}:& Forward Selection with $\xi=0$ and\\
\textbf{FS-1}:& Forward Selection with $\xi=1$ (see section~\ref{sec:met-fs}).\\
\end{tabular}
\medskip

%\noindent
Besides the comparison of feature selection methods, further interest lies in 
the analysis of strategies for the hyperparameter $\xi$. We compare a 
\textbf{cost-agnostic strategy} with a fixed value $\xi=0$, a 
\textbf{simple-BCR strategy} with a fixed value $\xi=1$ and a
\textbf{hyperparameter tuning strategy} with an adaptively optimized value 
$\xi$ based on Grid Search. An introduction of these strategies is given in
Subsection~\ref{sec:hyper}. For the real-world simulations, the cost-agnostic
versions define the baselines for each of the proposed methods, respectively.

\subsection{Performance Measure and Monte Carlo SE}
The MMCE of a Random Forest model (no individual parameter tuning:
\texttt{num.trees =} $1000$~,~\texttt{mtry =} $\sqrt{p}$) applied on 
independent test data is used to summarize and assess the selected feature 
sets. For a global comparison of all methods and strategies described in the
previous sections, the mean of the MMCEs over the 100 simulation runs is taken.
This value represents our final performance measure for each method and is
labeled $\overline{\text{MMCE}}$ in the following. As comparisons of this
measure can only be assessed in the context of the simulation setup, we 
consider the uncertainty in terms of the Monte Carlo standard error. An
approximate estimate of the Monte Carlo standard error (SE) of
$\overline{\text{MMCE}}$ is given by
\begin{equation} \label{eq:mcse}
  \text{Monte Carlo SE}(\overline{\text{MMCE}}) = 
  \sqrt{\frac{\widehat{\text{Var}}(\text{MMCE})}{n_\text{sim}}} .
\end{equation}
A two-sided $\alpha$-level confidence interval for $\overline{\text{MMCE}}$ can
thus be written as
\begin{equation} \label{eq:cimmce}
  \text{CI}_\alpha(\overline{\text{MMCE}}) : \left[\overline{\text{MMCE}} \pm 
  q_{1-\frac{\alpha}{2}} \cdot \text{Monte Carlo SE}(\overline{\text{MMCE}}) 
  \right] ,
\end{equation}
where $q_{1-\frac{\alpha}{2}}$ is the $(1-\frac{\alpha}{2})$-quantile of the 
standard normal distribution.

With regard to the individual run-times of all methods and the factorial design 
of the study, the definition of $n_\text{sim} = 100$ provides an acceptable 
reduction of uncertainty at a feasible computational effort. To account for 
this uncertainty, all following results comparing $\overline{\text{MMCE}}$ 
values are reported with their respective 95\%-confidence intervals as defined
in Equation~(\ref{eq:cimmce}). Additionally, all individual Monte Carlo 
standard errors for every setting, budget limit and method are tabulated and 
can be found in \nameref{add:1}.

%------------------------------------------------------------------------------%
\section{Results}\label{sec:res}
\subsection{Artificial Data Results}
The first analysis focuses on the proposed BCR criterion of each method. For 
all hyperparameter configurations, the MMCE on the test data set is computed. 
An exemplary overview of the results is given in 
Figure~\ref{fig:artificial-methods} for setting D with budget limit 10. This
setting was chosen as it shows many characteristics also found in the other
settings of the factorial design. Corresponding illustrations for every setup
can be found in Appendix~\ref{add:2}.

\begin{figure}[ht]
  \includegraphics[width=\textwidth]{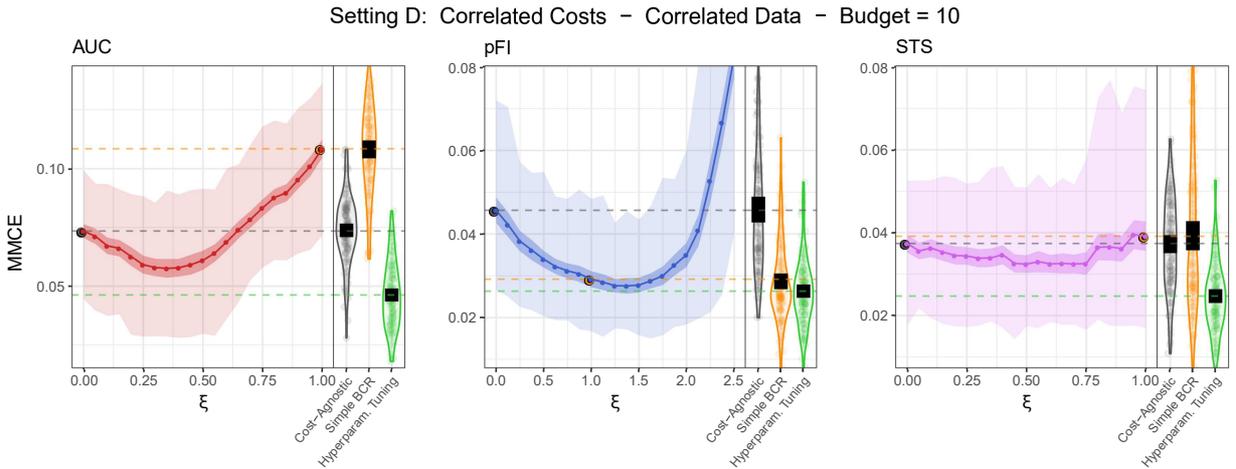}
  \caption{Three plots with two subplots each corresponding to the analyzed
  feature selection methods using the BCR criterion. Each left subplot shows a
  colored line of connected points representing the mean MMCE over all 100 
  simulation runs at a grid of $\xi$ values. The dark-shaded ribbon represents 
  a 95\%-CI around this mean value. The light shaded area shows the region 
  between the 5\% and 95\% quantiles of the empirical distribution of the MMCE. 
  The mean MMCE at $\xi = 0$ and $\xi = 1$ are highlighted with a gray and an 
  orange point, respectively. Each right subplot shows violin plots of the
  empirical MMCE distributions for the analyzed hyperparameter strategies. The
  95\%-CI region of the mean is given by a black box over the violins. Finally,
  the mean MMCE values of the three strategies are annotated with dashed lines
  over both subplots. Note that the hyperparameter tuning strategy (green) does
  not correspond to a single point on the left subplots, as $\xi$ is chosen at
  each simulation run independently of the OOB error of the current model. This
  plot shows results for setting D with $c_\text{max} = 10$. Corresponding 
  plots for all other levels of the factorial design are given in
  Appendix~\ref{add:2}\label{fig:strategies}} 
\end{figure}

All resulting curves of Figure~\ref{fig:strategies} are convex, i.e.\ 
with larger values of $\xi$ the MMCE values first decrease and after a certain
point begin to increase again, eventually exceeding the initial MMCE value at 
$\xi=0$. This trend is also observed in almost all other analyzed settings. The 
only exceptions to this is STS for setting~C, with $c_\text{max} = 2$ and 
$c_\text{max} = 5$, where the minimum occurs at $\xi=0$ and MMCE values 
increase from there on. When comparing the cost-agnostic and the simple BCR 
strategy, none is generally superior for all settings. In 
Figure~\ref{fig:strategies} the results already differ greatly between methods. 
The cost-agnostic strategy is superior for AUC, the simple BCR strategy is 
superior for pFI and both are approximately equal for STS. This uncertainty is 
also observed for other settings with no general preference for either of both 
strategies. Furthermore, the actual minima of the curves do not tend towards 
any fixed value, but also differ over all analyzed settings and methods. An 
alternative to defining a fixed $\xi$ is given by the hyperparameter tuning 
strategy, which selects a separate value of $\xi$ in each simulation run. With 
this adaptive strategy, the tuning approach is able to produce generally lower 
mean MMCE values than any point on the curves. In the majority of settings, 
hyperparameter tuning resulted in notably lower mean MMCE values compared 
to the cost-agnostic and simple BCR strategies. With larger budget values, 
differences between strategies decrease on an absolute scale, yet remain 
similar relative to the Monte Carlo SE of the MMCEs. The data and 
cost-correlation definitions of settings A to D showed no further effects. 

In all following method comparisons we focus on simulation results of the 
hyperparameter tuning strategy for AUC, pFI and STS and omit the inferior 
cost-agnostic and simple BCR strategies. To compare the proposed feature 
selection methods for every level of the factorial simulation design, a 
comprehensive illustration of all MMCE results is given in 
Figure~\ref{fig:artificial-methods}.

\begin{figure}[h]\centering
  \includegraphics[width=0.95\textwidth]{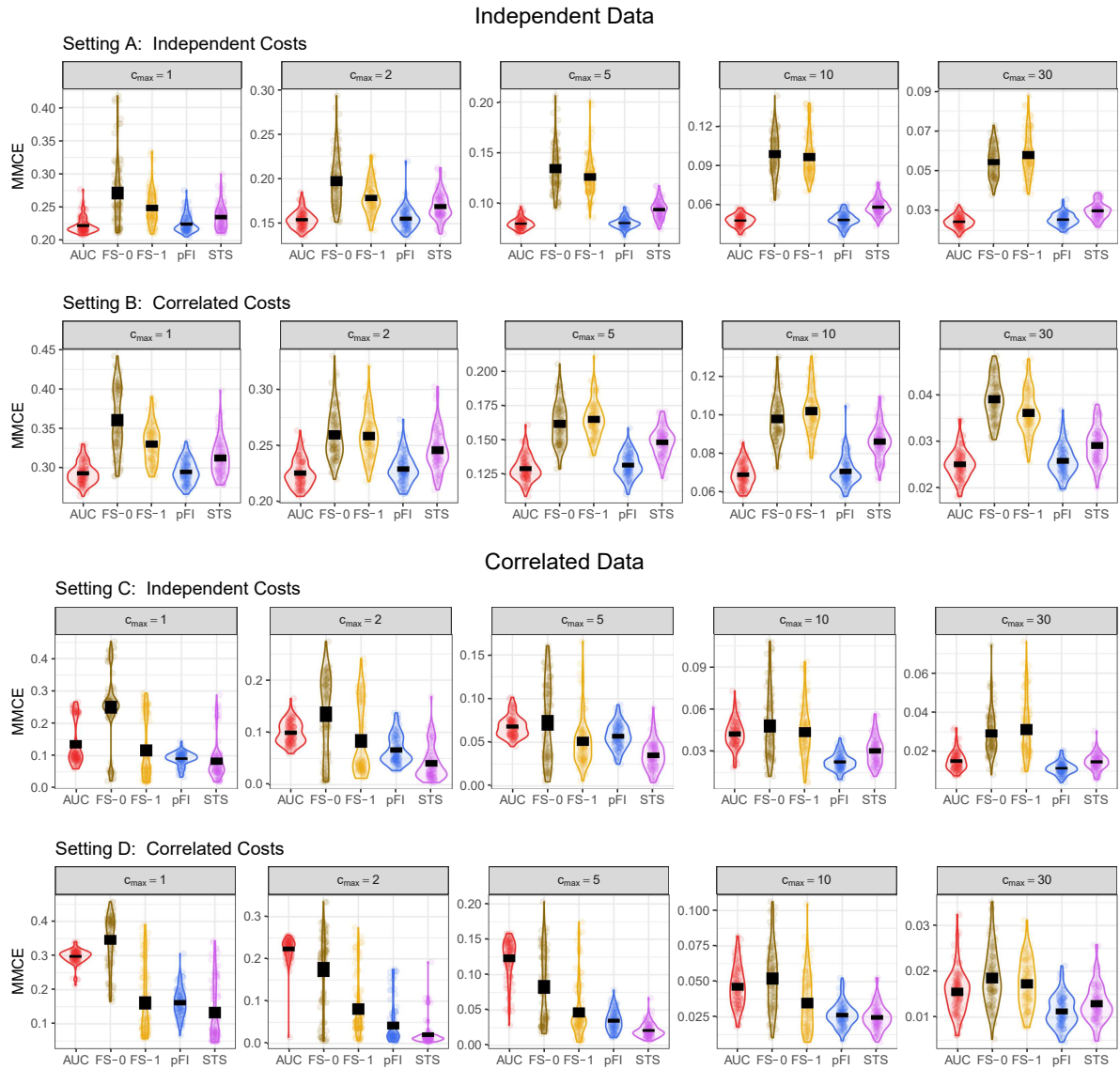}
  \caption{MMCE values for all simulation settings and budget limits of the 
  artificial data simulation. Violin plots for every feature selection method
  illustrate the empirical distributions of the MMCE values obtained in the 100 
  simulation runs (transparent dots). The 95\%-CI region of the mean MMCE of 
  each distribution is given by a black box over the 
  violins.\label{fig:artificial-methods}}
\end{figure}

The first two rows of this figure illustrate the independent data settings A 
and B. In these settings, the univariate filter method AUC shows the lowest 
mean MMCE closely followed by pFI. STS comes in third with significantly larger 
mean MMCE values. Here, and in the following, ``significant'' refers to the
(unadjusted) 95\%-Monte Carlos SE confidence intervals of the mean MMCEs. We
conservatively consider two methods to differ significantly, if two CIs do not
overlap. The worst performances are observed for FS-0 and FS-1, where similar 
to the results of the hyperparameter strategy analysis, none of both is
generally superior. For the correlated data settings C and D, the method
rankings differ notably. In most situations STS significantly outperforms the
alternative methods. Among the FS methods, FS-1 generally shows lower mean MMCE
values compared to FS-0 here. Yet, in almost all cases both cannot compete with
pFI and STS. The same applies to AUC, which also typically appears at the 
bottom end of the ranking. For all settings A to D, increasing the size of the
budget limit results in smaller absolute differences between methods. In
settings C and D, differences relative to the Monte Carlo SE of the MMCEs also
slightly decrease with higher budget limits. This effect is, however, not
notable in settings A and B. Correlation of feature relevance and costs 
revealed no obvious effects in our results. Altogether, the main influencing
factor of the performance ranking in our artificial simulation is if 
independent or correlated data are present.

\subsection{Real-World Data Results}
\begin{figure}[h]
  \centering
  \includegraphics[width=0.8\textwidth]{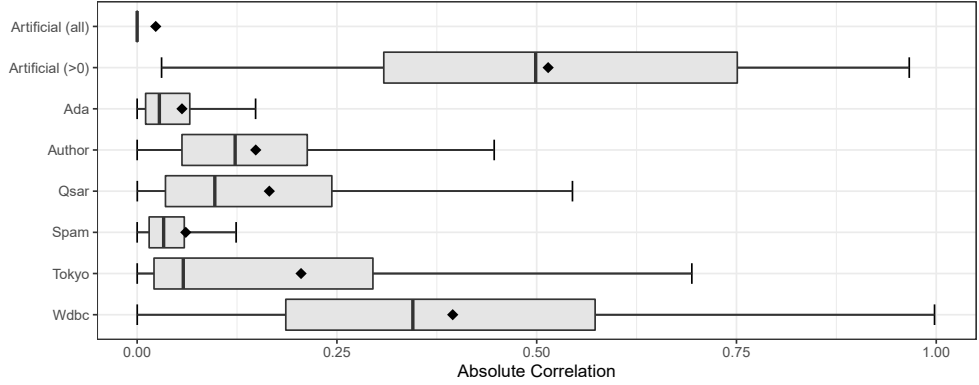}
  \caption{Boxplots for the absolute correlation values of all real-world data 
  sets. Additionally, correlation boxplots for the block-diagonal matrix of
  Equation~\ref{eq:blocksig} used in the artificial simulation are shown (first
  two lines), including also zeros (all) and including only non-zeros
  ($>0$).\label{fig:cor}}
\end{figure}

As the data correlation notably influenced the results of the artificial 
simulation, our first analysis of the real-world data concerns the correlation
between the features. Boxplots of the absolute correlation values observed in
the six real-world data sets and corresponding boxplots for the artificial
settings are given in Figure~\ref{fig:cor}. 

To highlight the design of the artificial simulations, the block-diagonal 
structure of $\boldsymbol\Sigma$ in Equation~\ref{eq:blocksig} is illustrated 
in two versions. The first version, which includes all values, shows that only 
very few actually correlated entries are found in $\boldsymbol\Sigma$. However, 
these non-zero entries cover a large range, which is illustrated by the second 
boxplot. The observed real-world correlations differ from this structure. Ada 
and Spam contain rather small correlations with an average around $0.1$. 
Compared to these, the boxplots of Author, Qsar and Tokyo include notably 
higher correlations and have their mean at approximately $0.2$. The highest
correlations are found at Wdbc, with multiple strong correlations above $0.9$
and a mean of approximately $0.4$. 

\begin{figure}[h]
  \includegraphics[width=\textwidth]{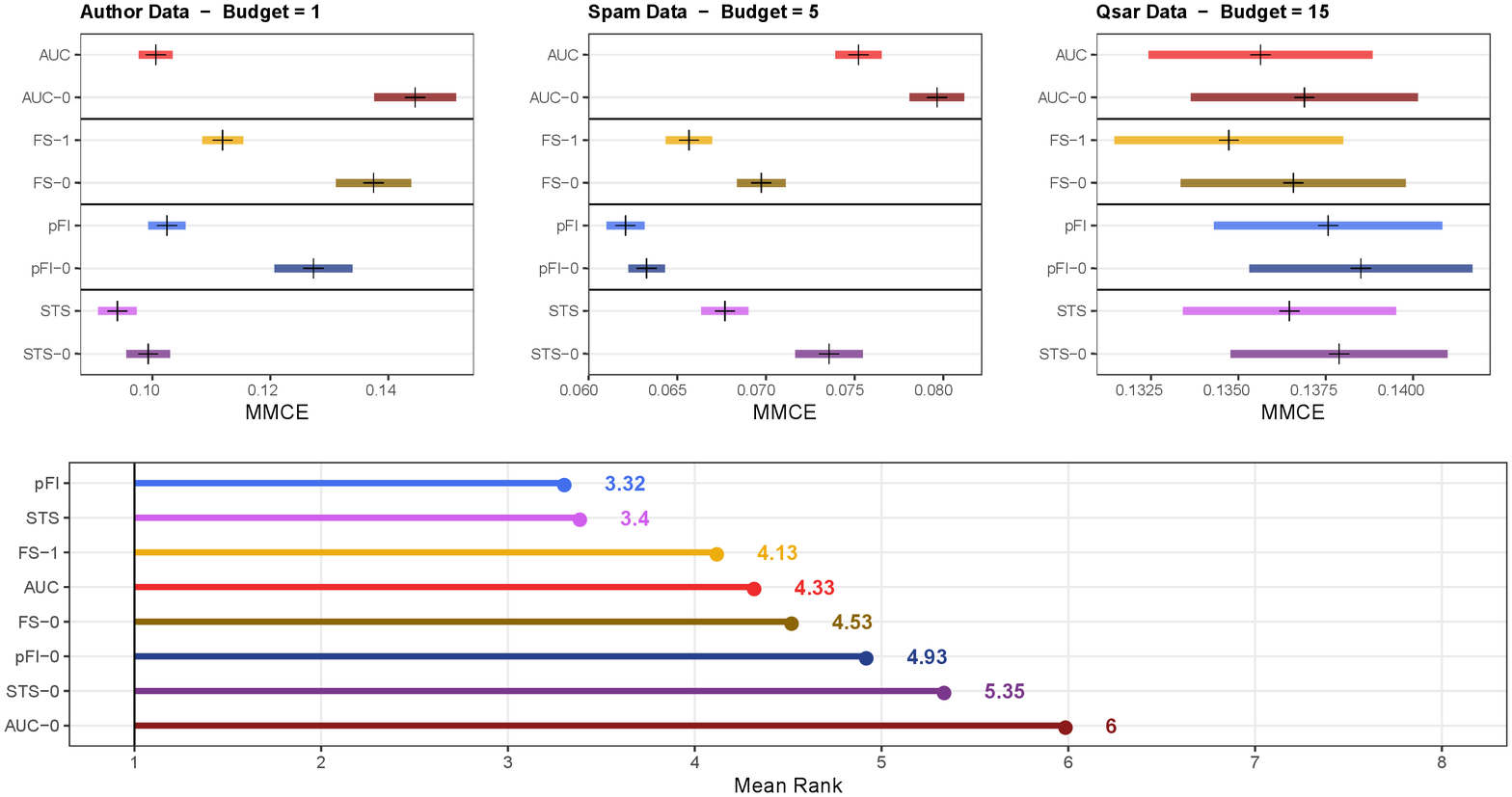}
  \caption{Top: Method comparison results of three representative real-world 
  data sets and budget limits $c_\text{max}$. Crosses show the mean MMCE for 
  each feature selection method, and horizontal bars show the corresponding
  95\%-Monte Carlo confidence intervals of these means. The cost-agnostic
  baseline version of each method is highlighted by adding ``-0'' to its name.
  Bottom: Mean rank of each method over all analyzed real-world simulation
  settings. Methods on the y-axis are sorted by mean 
  rank.\label{fig:realworld_methods}}
\end{figure}

The main aim of the real-world simulation is a comparison of the proposed
methods for every analyzed data set and budget limit. To illustrate these
results in a comprehensible format, we depict the main performance criterion
($\overline{\text{MMCE}}$) together with its respective 95\%-Monte Carlo CI for
all setups. A large plot-matrix of these results can be found in
Appendix~\ref{add:3}. An overview of the main results from representative 
setups is given in Figure~\ref{fig:realworld_methods}.

Altogether, the results vary strongly between different setups. Each 
non-baseline method significantly dominates the other methods in at least one 
of the setups. Examples of such ``best'' methods are: AUC for Qsar with
$c_\text{max} = 1$, FS-1 for Wdbc with $c_\text{max} = 1$, pFI for Spam with
$c_\text{max} = 5$, and STS for Author with $c_\text{max} = 1$. The latter two
are also illustrated in Figure~\ref{fig:realworld_methods}. Method rankings do
not only vary between different data sets, but also change for different budget
limits. For Spam, for example, the best method changes along four of the five
analyzed budget limits, see Appendix~\ref{add:3}. Similar to the artificial
data, differences between methods and the size of the CIs decrease with
increasing budget. For the Spam data, we also analyzed effects of different
random draws of the cost vector by sampling 5 versions of this vector. Detailed
plots of the individual results for each random draw can be found in 
Appendix~\ref{add:4}. While a general tendency to favor pFI can be observed 
for most Spam data versions, the analysis shows that different rankings can 
also arise from different explicit cost setups. Comparing the results with
respect to the correlations shown in Figure~\ref{fig:cor}, no notable
associations can be observed. For Qsar, which has relatively high correlations,
the univariate AUC method performed best. On the contrary, the relatively weak
correlations of Ada favored more multivariate-oriented methods like FS and STS.
Both showed weaknesses for non-correlated data in the artificial simulations.
When categorizing the data sets by their field according to 
Table~\ref{tab:realsets}, no obvious groups can be found as well. Author and 
Spam, for instance, both originating from a similarly structured text mining 
application, both benefit from completely different methods. While individual
rankings may vary for different setups, a clear overall preference for
non-baseline methods is notable in almost all setups.

To obtain a global measure of the simulation results, the average rankings of
each method over all analyzed data sets (excluding the cost-vector repetitions 
of Spam) are illustrated in the bottom plot of 
Figure~\ref{fig:realworld_methods}. This plot shows that on average the whole
group of baseline methods ranks worse than all non-baseline methods. Within 
the non-baseline methods, STS and pFI furthermore show an improvement of
almost a full rank on average. Altogether, while local differences highlight
that no universally superior method exists for all setups, the global 
rankings show notable preferences towards non-baseline approaches, specifically
STS and pFI.

%------------------------------------------------------------------------------%
\section{Discussion}\label{sec:dis}
With the results presented in the previous section, we address all aims defined 
in Section~\ref{sec:aims}, as follows. Aim one concerns the hyperparameter
in the BCR criterion (see Section~\ref{sec:hyper}) and compares the proposed 
cost-agnostic and simple BCR strategies. The results in 
Figure~\ref{fig:strategies} show that there is no generally superior strategy 
with a fixed value for $\xi$. That means, not only the proposed strategies, but 
also other possible choices for $\xi$ are unsuited as a general solution for 
different data sets or even different budget limits within a single data set.

Aim two extends this topic by evaluating the benefits of a hyperparameter 
tuning strategy. In almost all simulations, this strategy notably improved the 
overall feature selection results. By not defining a global value $\xi$, but 
selecting the optimal value for each simulation run independently, 
hyperparameter tuning outperformed any fixed $\xi$ approach. As the additional 
computational complexity does typically not create run-time issues for fast 
methods like AUC, pFI or STS, we therefore generally recommend to use 
hyperparameter tuning for these approaches.

The third aim analyzes strengths and weaknesses of methods on univariate and
multivariate data sets. The univariate case corresponds to the results of the
artificial settings A and B. Here, AUC generally provides strong results. This
seems reasonable, as the theoretical idea of a univariate filter exactly 
targets the simulated data scenario. STS, which focuses on multivariate tree
structures, shows some weakness at purely univariate setups. While FS is also
notably inferior here, we avoid general conclusions on this method, as it uses
the inferior cost-agnostic and simple BCR strategies. pFI handles the data
structure similarly to AUC and can thus also be considered well suited for
univariate settings.

The multivariate data case corresponds to the artificial settings C and D. 
Here, STS outperforms the other methods in almost every setup. While the
multivariate nature of these settings emphasizes the strengths of STS, it also
exposes the main weakness of AUC, which as a univariate filter approach is not
able to identify the relevant structures and clearly falls behind here. pFI
ranks features by their importance in a Random Forest fitted on all features.
The multivariate aspect of this approach renders it superior to AUC. However,
for most of the analyzed setups, filtering by this importance measure is
inferior to the procedure of STS, which evaluates smaller, but coherent
combinations. Nevertheless, pFI still provides the second best alternative 
here and can be considered well-suited as well. Method rankings of FS are
inconsistent and vary between setups.

Aim four concerns effects of different budget limits $c_\text{max}$, and 
possible influences from a correlation of true feature effect sizes and 
feature costs. In some artificial and real-world settings, larger budgets 
reduce the differences between approaches and the actual method choice becomes
less relevant. This does however not hold true in general. In many real-world
data sets, the best method changes for different budget limits. Nevertheless,
there is no notable systematic preference of methods for small or large 
budgets. Besides these mentioned aspects, no further budget effects are 
observed. The same applies to the analyses on correlated effect sizes and 
feature costs. While this setup defines a realistic scenario, which in most
simulations influences the selection of the features, it does not affect the 
method comparison results in any major way.  

The fifth aim addresses further effects detected on real-world data that extend 
the findings of the artificial simulations. From the real-world results we 
conclude that our artificial design is only able to tackle a few aspects of the
actual data generating process of many real-world data sets. The individuality
of the results hints towards many unknown factors, which further influence the
suitability of a method. However, these aspects cannot be specified from the
data at hand and would require further detailed research.

Aim six focuses on the comparison to the baseline methods in the real-world
analysis. Results of nearly all individual setups show that the baseline 
versions of all methods are inferior to their non-baseline counterparts. A 
global analysis of the average rankings of each method furthermore highlighted
that even the lowest ranking non-baseline method (AUC) ranks above the best
baseline method (FS-0). Therefore, we conclude that the proposed methods
outperformed the cost-agnostic baseline approaches. 

The final aim considers categorizing method preferences of real-world data sets 
according to given Metadata. We could not identify common factors of the 
real-world data sets that lead to a preferred method. Moreover, the results 
show evidence that local differences in the highest ranking method can already
arise from different budget limits or explicit cost definitions. On a global
level, however, an analysis of average method rankings shows that STS and pFI
are more versatile than other methods and on average rank notably higher.

%------------------------------------------------------------------------------%
\section{Conclusion and Outlook}\label{sec:out}
Our paper proposed multiple methods for cost-sensitive feature selection
in the context of Random Forests. The first of these is a novel approach named
STS. STS identifies a set of relevant small tree structures from an ensemble of
candidate trees and selects the minimal feature subset that is able to build
these structures. It thus provides a computationally efficient solution for a
multivariate feature selection problem. Besides STS, we also proposed three
further adaptations of common feature selection approaches for Random Forests.
With AUC as a univariate filter approach, pFI as a semi-multivariate approach 
and FS as a true multivariate, but computationally complex approach, a diverse
set of methods was defined for the problem at hand. In a large scale simulation
study of four artificial data settings and six real-world data sets, strengths
and weaknesses of all methods were assessed. There is no universal one-fits-all
method among our candidates. Each approach was superior in at least one of the
analyzed settings. This illustrates the fundamental differences in the basic
selection strategies of each algorithm, which can be more or less suited in
different data situations. The factors guiding the optimality of a method may,
however, often depend on possibly unknown data generating mechanisms. For each
method, an individual use-case setting could be found in our simulation. A 
global analysis of the rankings of each method over \textit{all} data sets
nevertheless showed that STS and pFI on average ranked notably higher than 
their competitors and may thus be considered favorable candidates in unknown
situations. Nevertheless, for a practical problem, we still recommend applying
multiple methods (e.g. the computationally fast AUC, pFI and STS approaches) 
and base a final decision on a comparative analysis. If the high complexity of 
FS is manageable for the problem at hand, we recommend to also include this
approach. Regardless of the method, all BCR approaches discussed in this paper
highly benefited from a hyperparameter tuning, and we therefore finally advise
to ensure an adequate trade-off of performance and cost with a
hyperparameterized BCR criterion.

Beyond the developments in this paper, in the following we mention a few 
further ideas for future research on this topic. In the current implementation 
of STS, the number of features per tree is limited by the tree depth. This is 
a practical solution, which has multiple downsides. First, a limited depth
generally reduces predictive performance. Second, this approach does not
guarantee that exactly the intended number of different features per tree is
selected. An alternative to limiting the tree depth could be to limit the 
number of features a tree may use instead and allow the tree to fully grow. 
For example, if we intend to build a three-feature tree, the tree-growing
algorithm may perform normal splits until three distinct features have been
used. From then on, it may continue to perform further splits, but only with
these three features. Instead of limiting the number of features, limiting the
cost of the tree itself might be another viable option. This would extend the
set of candidate trees even further. Moreover, trees could also be generated 
in a cost-sensitive manner that includes a trade-off between costs and
performance at each split. 

In the STS method described in this work, the base ensemble of trees is 
generated from a Random Forest framework. However, it should be noted that any 
tree growing algorithm generating heterogeneous trees may be used instead. With 
a more general approach, redundant trees could be avoided or special 
multivariate structures could be emphasized. 

For future analyses, alternative strategies for the hyperparameter tuning --
apart from a simple Grid Search -- might increase performance and reduce the 
run-time of all methods. Especially for FS, such developments may be a relevant 
aspect deciding if the method is feasible at all. 
Finally, all current simulations specialize on binary classification. Yet none 
of the proposed methods is technically limited to this setup. Analyzing the 
effects on different response and model types therefore also provides a good 
basis for further research in this field.

%------------------------------------------------------------------------------%
\section*{Appendix}
\subsection*{Additional file 1}\label{add:1}
\textit{Table of all Monte Carlo standard errors. (PDF)}
\smallskip

\noindent
Monte Carlo SE estimates of the mean MMCE after $n_\text{sim} = 100$ simulation 
runs. The table shows estimates for all analyzed methods, budget limits and  
data sets of the simulation study of this paper.

\subsection*{Additional file 2}\label{add:2} 
\textit{Comprehensive version of Figure~\ref{fig:strategies} including all 
levels of the factorial design for the artificial data simulation. (PDF)}
\smallskip

\noindent
Each full figure illustrates a certain setting and budget limit combination.
Details on the individual elements of these figures can be found in the 
description of the analogously structured Figure~\ref{fig:strategies}.

\subsection*{Additional file 3}\label{add:3} 
\textit{Plot matrix of all analyzed real-world data sets (rows) and budget 
limits $c_\text{max}$ (columns). (PDF)}
\smallskip

\noindent
Crosses show the mean MMCE for each feature selection method. Horizontal bars
show the corresponding 95\%-Monte Carlo confidence intervals of these means.
Methods on the y-axis are given in increasing order of mean MMCE to provide a
visual ranking for each simulation setup. The cost-agnostic baseline version 
of each method is highlighted by adding ``-0'' to its name.

\subsection*{Additional file 4}\label{add:4} 
\textit{Plot matrix of the Spam data results for five random draws of the 
cost vector. (PDF)}
\smallskip

\noindent
Crosses show the mean MMCE for each feature selection method. Horizontal bars
show the corresponding 95\%-Monte Carlo confidence intervals of these means.
Methods on the y-axis are given in increasing order of mean MMCE to provide a
visual ranking for each simulation setup. The cost-agnostic baseline version 
of each method is highlighted by adding ``-0'' to its name.

%\begin{acknowledgements}
%\end{acknowledgements}


\section*{Declarations}
\subsection*{Funding}
This work was supported by Deutsche Forschungsgemeinschaft (DFG), Project
RA 870/7-1, and Collaborative Research Center SFB 876, A3. The authors
acknowledge financial support by Deutsche Forschungsgemeinschaft and
Technische Universität Dortmund within the funding programme Open
Access Publishing. 

\subsection*{Conflict of interest}
The authors declare that they have no conflict of interest.

\subsection*{Availability of data and material}
All data sets and material used in this paper are available on Github:
\url{https://github.com/RudolfJagdhuber/STS}
% This repository is private at the moment and will be set to public as 
% soon as the paper is published

\subsection*{Code availability}
The full code used in this paper is available on Github: 
\url{https://github.com/RudolfJagdhuber/STS}

\subsection*{Authors contributions}
Rudolf Jagdhuber, developed all methods, designed and executed the simulation 
studies, interpreted the results, and wrote the manuscript.
Michel Lang proposed the first idea of Tree Selection and corrected and 
approved the manuscript.
Jörg Rahnenführer initiated the topic, supervised the project, contributed to 
the problem definition, the design of the simulation study and to the 
interpretation of the results, and corrected and approved the manuscript.

\bibliographystyle{spbasic}      % basic style, author-year citations
\bibliography{Manuscript_wFigures}

\begin{thebibliography}{30}
\providecommand{\natexlab}[1]{#1}
\providecommand{\url}[1]{{#1}}
\providecommand{\urlprefix}{URL }
\expandafter\ifx\csname urlstyle\endcsname\relax
  \providecommand{\doi}[1]{DOI~\discretionary{}{}{}#1}\else
  \providecommand{\doi}{DOI~\discretionary{}{}{}\begingroup
  \urlstyle{rm}\Url}\fi
\providecommand{\eprint}[2][]{\url{#2}}

\bibitem[{Banas et~al.(2018)Banas, Neumann, Eiglsperger, Schiffer, Putz,
  Reichelt-Wurm, Kr{\"a}mer, Pagel, and Banas}]{banas2018identification}
Banas M, Neumann S, Eiglsperger J, Schiffer E, Putz FJ, Reichelt-Wurm S,
  Kr{\"a}mer BK, Pagel P, Banas B (2018) Identification of a urine metabolite
  constellation characteristic for kidney allograft rejection. Metabolomics
  14(9):116

\bibitem[{Bischl et~al.(2016)Bischl, Lang, Kotthoff, Schiffner, Richter,
  Studerus, Casalicchio, and Jones}]{bischl2016mlr}
Bischl B, Lang M, Kotthoff L, Schiffner J, Richter J, Studerus E, Casalicchio
  G, Jones ZM (2016) mlr: Machine learning in r. Journal of Machine Learning
  Research 17(170):1--5,
  \urlprefix\url{https://mlr.mlr-org.com/articles/tutorial/tune.html}

\bibitem[{Bischl et~al.(2017)Bischl, Richter, Bossek, Horn, Thomas, and
  Lang}]{bischl2017mlrmbo}
Bischl B, Richter J, Bossek J, Horn D, Thomas J, Lang M (2017) mlrmbo: A
  modular framework for model-based optimization of expensive black-box
  functions. arXiv preprint arXiv:170303373

\bibitem[{Bol{\'o}n-Canedo et~al.(2014)Bol{\'o}n-Canedo, Porto-D{\'\i}az,
  S{\'a}nchez-Maro{\~n}o, and Alonso-Betanzos}]{bolon2014framework}
Bol{\'o}n-Canedo V, Porto-D{\'\i}az I, S{\'a}nchez-Maro{\~n}o N,
  Alonso-Betanzos A (2014) A framework for cost-based feature selection.
  Pattern Recognition 47(7):2481--2489

\bibitem[{Bommert et~al.(2020)Bommert, Sun, Bischl, Rahnenf{\"u}hrer, and
  Lang}]{bommert2020benchmark}
Bommert A, Sun X, Bischl B, Rahnenf{\"u}hrer J, Lang M (2020) Benchmark for
  filter methods for feature selection in high-dimensional classification data.
  Computational Statistics \& Data Analysis 143:106839

\bibitem[{Boulesteix et~al.(2017)Boulesteix, De~Bin, Jiang, and
  Fuchs}]{boulesteix2017ipf}
Boulesteix AL, De~Bin R, Jiang X, Fuchs M (2017) Ipf-lasso:
  Integrative-penalized regression with penalty factors for prediction based on
  multi-omics data. Computational and mathematical methods in medicine 2017

\bibitem[{Breiman(2001)}]{breiman2001random}
Breiman L (2001) Random forests. Machine learning 45(1):5--32

\bibitem[{De~Meyer et~al.(2008)De~Meyer, Sinnaeve, Van~Gasse, Tsiporkova,
  Rietzschel, De~Buyzere, Gillebert, Bekaert, Martins, and
  Van~Criekinge}]{de2008nmr}
De~Meyer T, Sinnaeve D, Van~Gasse B, Tsiporkova E, Rietzschel ER, De~Buyzere
  ML, Gillebert TC, Bekaert S, Martins JC, Van~Criekinge W (2008) Nmr-based
  characterization of metabolic alterations in hypertension using an adaptive,
  intelligent binning algorithm. Analytical chemistry 80(10):3783--3790

\bibitem[{Dua and Graff(2017)}]{dua2019}
Dua D, Graff C (2017) {UCI} machine learning repository.
  \urlprefix\url{http://archive.ics.uci.edu/ml}

\bibitem[{Fern{\'a}ndez-Delgado et~al.(2014)Fern{\'a}ndez-Delgado, Cernadas,
  Barro, and Amorim}]{fernandez2014we}
Fern{\'a}ndez-Delgado M, Cernadas E, Barro S, Amorim D (2014) Do we need
  hundreds of classifiers to solve real world classification problems? The
  journal of machine learning research 15(1):3133--3181

\bibitem[{Goschenhofer(2017)}]{openMLada}
Goschenhofer J (2017) Openml ada{\_}agnostic.
  \url{https://www.openml.org/d/40993}, accessed: 2020-05-01

\bibitem[{Guyon and Elisseeff(2003)}]{guyon2003introduction}
Guyon I, Elisseeff A (2003) An introduction to variable and feature selection.
  Journal of machine learning research 3(Mar):1157--1182

\bibitem[{Holland(1973)}]{holland1973genetic}
Holland JH (1973) Genetic algorithms and the optimal allocation of trials. SIAM
  Journal on Computing 2(2):88--105

\bibitem[{Jagdhuber et~al.(2020)Jagdhuber, Lang, Stenzl, Neuhaus, and
  Rahnenf{\"u}hrer}]{jagdhuber2020cost}
Jagdhuber R, Lang M, Stenzl A, Neuhaus J, Rahnenf{\"u}hrer J (2020)
  Cost-constrained feature selection in binary classification: adaptations for
  greedy forward selection and genetic algorithms. BMC bioinformatics
  21(1):1--21

\bibitem[{Jones et~al.(1998)Jones, Schonlau, and Welch}]{jones1998efficient}
Jones DR, Schonlau M, Welch WJ (1998) Efficient global optimization of
  expensive black-box functions. Journal of Global optimization 13(4):455--492

\bibitem[{Leskovec et~al.(2007)Leskovec, Krause, Guestrin, Faloutsos,
  Faloutsos, VanBriesen, and Glance}]{leskovec2007cost}
Leskovec J, Krause A, Guestrin C, Faloutsos C, Faloutsos C, VanBriesen J,
  Glance N (2007) Cost-effective outbreak detection in networks. In:
  Proceedings of the 13th ACM SIGKDD international conference on Knowledge
  discovery and data mining, ACM, pp 420--429

\bibitem[{Li et~al.(2014)Li, Min, and Zhu}]{li2014fast}
Li J, Min F, Zhu W (2014) Fast randomized algorithm for minimal test cost
  attribute reduction. International Journal of Reliability, Quality and Safety
  Engineering 21(06):1450028

\bibitem[{Li et~al.(2015)Li, Zhao, and Zhu}]{li2015fast}
Li J, Zhao H, Zhu W (2015) Fast randomized algorithm with restart strategy for
  minimal test cost feature selection. International Journal of Machine
  Learning and Cybernetics 6(3):435--442

\bibitem[{Mansouri et~al.(2013)Mansouri, Ringsted, Ballabio, Todeschini, and
  Consonni}]{mansouri2013quantitative}
Mansouri K, Ringsted T, Ballabio D, Todeschini R, Consonni V (2013)
  Quantitative structure--activity relationship models for ready
  biodegradability of chemicals. Journal of chemical information and modeling
  53(4):867--878

\bibitem[{Min and Xu(2016)}]{min2016semi}
Min F, Xu J (2016) Semi-greedy heuristics for feature selection with test cost
  constraints. Granular Computing 1(3):199--211

\bibitem[{Min et~al.(2011)Min, He, Qian, and Zhu}]{min2011test}
Min F, He H, Qian Y, Zhu W (2011) Test-cost-sensitive attribute reduction.
  Information Sciences 181(22):4928--4942

\bibitem[{Min et~al.(2014)Min, Hu, and Zhu}]{min2014feature}
Min F, Hu Q, Zhu W (2014) Feature selection with test cost constraint.
  International Journal of Approximate Reasoning 55(1):167--179

\bibitem[{Morris et~al.(2019)Morris, White, and Crowther}]{morris2019using}
Morris TP, White IR, Crowther MJ (2019) Using simulation studies to evaluate
  statistical methods. Statistics in medicine 38(11):2074--2102

\bibitem[{Niculescu-Mizil and Caruana(2005)}]{niculescu2005predicting}
Niculescu-Mizil A, Caruana R (2005) Predicting good probabilities with
  supervised learning. In: Proceedings of the 22nd international conference on
  Machine learning, pp 625--632

\bibitem[{Pacl{\'\i}k et~al.(2002)Pacl{\'\i}k, Duin, van Kempen, and
  Kohlus}]{paclik2002feature}
Pacl{\'\i}k P, Duin RP, van Kempen GM, Kohlus R (2002) On feature selection
  with measurement cost and grouped features. In: Joint IAPR International
  Workshops on Statistical Techniques in Pattern Recognition (SPR) and
  Structural and Syntactic Pattern Recognition (SSPR), Springer, pp 461--469

\bibitem[{Simonoff(2013)}]{simonoff2013analyzing}
Simonoff JS (2013) Analyzing categorical data. Springer Science \& Business
  Media

\bibitem[{Tan(1993)}]{tan1993cost}
Tan M (1993) Cost-sensitive learning of classification knowledge and its
  applications in robotics. Machine Learning 13(1):7--33

\bibitem[{Vanschoren et~al.(2013)Vanschoren, van Rijn, Bischl, and
  Torgo}]{OpenML2013}
Vanschoren J, van Rijn JN, Bischl B, Torgo L (2013) Openml: Networked science
  in machine learning. SIGKDD Explorations 15(2):49--60,
  \doi{10.1145/2641190.2641198},
  \urlprefix\url{http://doi.acm.org/10.1145/2641190.2641198}

\bibitem[{Wright and Ziegler(2015)}]{wright2015ranger}
Wright MN, Ziegler A (2015) ranger: A fast implementation of random forests for
  high dimensional data in c++ and r. arXiv preprint arXiv:150804409

\bibitem[{Zhou et~al.(2016)Zhou, Zhou, and Li}]{zhou2016cost}
Zhou Q, Zhou H, Li T (2016) Cost-sensitive feature selection using random
  forest: Selecting low-cost subsets of informative features. Knowledge-Based
  Systems 95:1--11

\end{thebibliography}

% \includepdf[pages=-, pagecommand={\vspace*{-3.5cm}\hspace*{-3.1cm}
% \fboxsep=5pt\fboxrule=1pt \fcolorbox{black}{white}{Additional\_file\_1.pdf}
% \thispagestyle{empty}},fitpaper=true]{Additional_file_1.pdf}

% \includepdf[pages=-, pagecommand={\vspace*{-3.5cm}\hspace*{-3.1cm}
% \fboxsep=5pt\fboxrule=1pt \fcolorbox{black}{white}{Additional\_file\_2.pdf}
% \thispagestyle{empty}},fitpaper=true]{Additional_file_2.pdf}

% \includepdf[pages=-, pagecommand={\vspace*{-3.5cm}\hspace*{-3.1cm}
% \fboxsep=5pt\fboxrule=1pt \fcolorbox{black}{white}{Additional\_file\_3.pdf}
% \thispagestyle{empty}},fitpaper=true]{Additional_file_3.pdf}

% \includepdf[pages=-, pagecommand={\vspace*{-3.5cm}\hspace*{-3.1cm}
% \fboxsep=5pt\fboxrule=1pt \fcolorbox{black}{white}{Additional\_file\_4.pdf}
% \thispagestyle{empty}},fitpaper=true]{Additional_file_4.pdf}

\end{document}